\RequirePackage[T1]{fontenc}
\documentclass[journal,letterpaper]{IEEEtran}
\usepackage{amsmath,amssymb,graphicx,booktabs,array,url,xcolor}
\usepackage[hidelinks]{hyperref}
\newcommand{\score}{SCORE}
\newcommand{\lm}{SCORE-LM}

\title{SCORE-LM: State-Space Radar Representations with Language Models for Fault Diagnosis}
\author{Mainak~Mallick and Seung-Kyum~Choi\thanks{The authors are with the Georgia Institute of Technology, Atlanta, GA, USA.}}
\hypersetup{pdfauthor={Mainak Mallick and Seung-Kyum Choi},pdftitle={SCORE-LM: State-Space Radar Representations with Language Models for Fault Diagnosis}}
\begin{document}
\maketitle
\begin{abstract}
Radar hardware faults threaten automated perception, motivating accurate, compact diagnosis and understandable maintenance guidance. We introduce SCORE-LM, which couples a small scatterer-conditioned operator-response encoder (SCORE) to an adapted local language model. SCORE combines self-referenced complex trajectories, physical descriptors, and a selective state-space branch, with source-only self-supervision and directional fault inference. On eight capture-excluded Rad-R fault recordings, it achieves state-of-the-art performance within the evaluated nine-model comparison: 88.39\% mean capture recall and 88.20\% four-fault macro-F1 at ten frames. Its 39,520 radar inference coefficients are 119.7$\times$ fewer than RadrNet-DS-CI's, while recall is 15.56 percentage points higher than this strongest competitor. In a separate low-label protocol, SCORE reaches 71.58\% recall with one labeled source window per class. A nonlinear projector converts four frozen fault similarities into five soft tokens, linking compact diagnosis to class-conditioned maintenance guidance. On 75 development questions covering 24 radar windows, language adaptation raises correct-fault answers from 45 to 62 (60.0\% to 82.7\%) relative to removing the co-trained adapters, while retaining the same projector. SCORE-LM thus combines a compact radar specialist with a language interface for communicating fault-specific inspection guidance.
\end{abstract}

\noindent\textbf{Note to Practitioners---}Automated vehicles and robotic platforms rely on radar measurements even when a sensor's mounting, protective cover, or receiver response has deteriorated. The proposed system first classifies a known fault family and then presents inspection-oriented guidance in natural language. Its radar encoder is small enough to separate continuous signal processing from an optional, larger language interface. The present implementation recognizes vibration, misalignment, blockage, and receiver degradation; its ten-frame decision rule assumes a fault is present. The language model translates class-similarity scores into fault-specific inspection suggestions. Guidance is intended to help an operator organize inspection and follow-up measurements, with equipment-specific procedures taking precedence. Experiments use a single-session research dataset. Deployment would require independent healthy recordings, testing across vehicles and operating conditions, and expert review of the recommendations. The design allows the classifier to remain available if the language component is omitted, and keeps diagnostic accuracy separate from the fluency of its explanation.


\section{Introduction}
Radar is a useful sensing modality for automation because it provides range and motion information without relying on visible illumination. However, a plausible measurement stream does not guarantee a healthy instrument. A displaced mounting, vibrating enclosure, obstructed radome, or changed receiver response can corrupt the signal before downstream perception operates. Hardware diagnosis therefore asks a different question from scene recognition: does the measured response resemble a known fault of the sensing system?

Two obstacles make this problem difficult. First, radar observations depend on the environment as well as the instrument. A classifier may associate a particular scene or recording with a fault label without learning transferable fault evidence. Second, obtaining fault labels requires controlled hardware interventions, making large and diverse training sets difficult to assemble. Rad-R provides a relevant starting point: raw-ADC recordings of four controlled hardware fault families and synchronized auxiliary sensors, acquired with a cascaded automotive radar~\cite{radr}. Its distinct recordings permit a capture-excluded evaluation, although the available data do not separate severity changes from recording identity.

An additional obstacle appears after classification. An operator may need an understandable statement of the condition and appropriate follow-up checks, not only an integer label. Language models offer a flexible interface, but fluent responses can add observations that were never measured. A model that correctly says ``blockage'' can still be unreliable if it invents a complete absence of echoes, an affected receiver, or a verified repair outcome. Classification accuracy and answer reliability must therefore be evaluated separately.

We study \lm{}, a modular connection between a radar specialist and a local language model. Its radar component, \score{}, represents a frame as a set of scatterer-conditioned responses. A shared two-branch encoder combines a selective state-space model (SSM) of complex loop trajectories with a nonlinear physical-descriptor branch. Uniform pooling summarizes peaks and frames. Self-supervised source training is followed by a directional fault readout. For the language stage, the radar encoder and readout are frozen. Four real-valued similarities, rather than a textual predicted label, condition a trainable multilayer projector and low-rank language adapters.

The language task is fault-conditioned maintenance guidance. A learned projector maps four diagnostic scores into the language model's embedding space. BearLLM established a precedent for connecting a diagnostic network to a language interface~\cite{bearllm}; our study develops this approach for automotive radar.

Our contributions are threefold. First, we specify a response-centered radar encoder and source-only learning protocol, including its fixed signal processing. Second, we compare fault recognition, label efficiency, observation budget, and selected processing costs against eight independently implemented or author-supplied learned methods. Third, we examine a score-to-language interface with controlled input/projector variants and a language-adapter removal control, distinguishing fault correctness, agreement with the specialist, and unsupported claims.

\section{Related Work and Scope}
\subsection{Radar hardware diagnosis}
The Rad-R benchmark introduces controlled vibration, yaw misalignment, radome blockage, and receive-channel degradation in a four-chip radar cascade~\cite{radr}. Its RadrNet family includes raw-IQ state-space and range--Doppler branches. RadrNet-DS-CI replaces absolute-phase processing with capture-invariant preprocessing and combines IQ and RD features through an adaptive gate. We evaluate the author-supplied DS-CI architecture under our eight-fold protocol; the original benchmark uses two cross-severity folds.

Our other competitors cover supervised RD learning and domain-generalization regularization. Deep CORAL aligns feature covariances~\cite{coral}; our source-only adaptation uses source severity groups. MixStyle mixes feature statistics during training~\cite{mixstyle}. DARM uses distance-aware instance/prototype objectives~\cite{darm}, while DDDG combines dual disentanglement with distribution diversification~\cite{dddg}. We adapt these machinery-fault methods to radar using independent encoders.

Domain-generalization surveys distinguish learning from available source conditions from adaptation using target data~\cite{dgsurvey,faultsurvey}. Recent machinery-fault approaches include semantics-consistent representation learning (SCRL)~\cite{scrl}, balancing discrepancy and consistency (BDC)~\cite{bdc}, and a discrete-wavelet convolutional network with cross-contrast perturbation (DWCN)~\cite{dwcn}. Our radar adaptations of BDC and DWCN extend the comparison with adversarial discrepancy learning and wavelet-based cross-contrast perturbation.

DomainBed identifies model selection and inconsistent experimental conditions as important obstacles to domain-generalization comparisons~\cite{domainbed}. We therefore specify target exclusion and final-checkpoint selection, and distinguish comparisons of complete recipes from controlled component interventions.

\begin{figure*}[!t]
\centering\includegraphics[width=\textwidth]{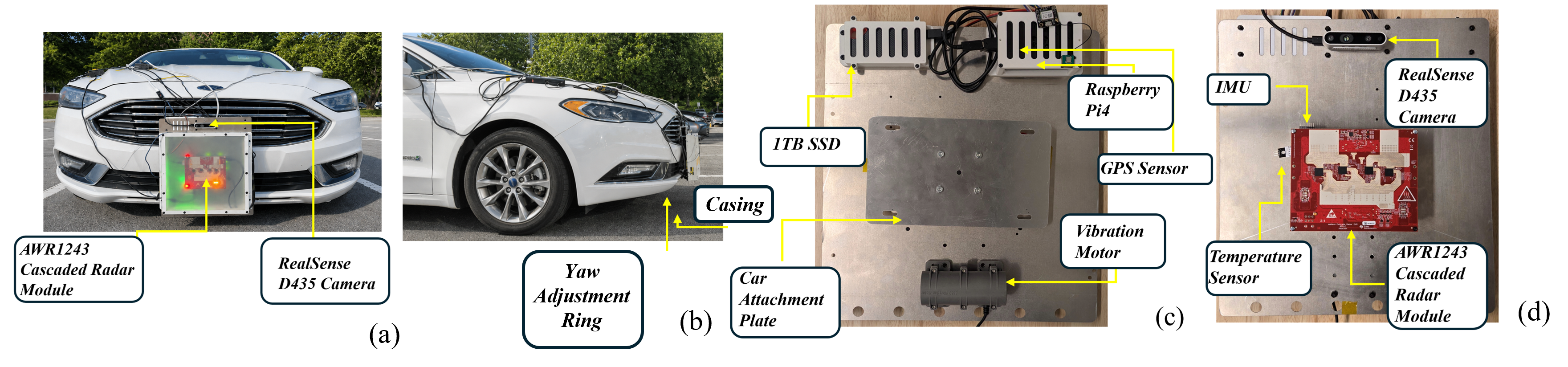}
\caption{Rad-R capture setup, arranged from the authors' dataset assets~\cite{radr}. From left: front and side vehicle views, followed by the two sensor-plate views. The hardware includes the cascade radar, camera, inertial and environmental sensors, positioning, storage, and controlled-vibration hardware.}
\label{fig:rig}
\end{figure*}

\begin{figure}[!t]
\centering\includegraphics[width=\columnwidth]{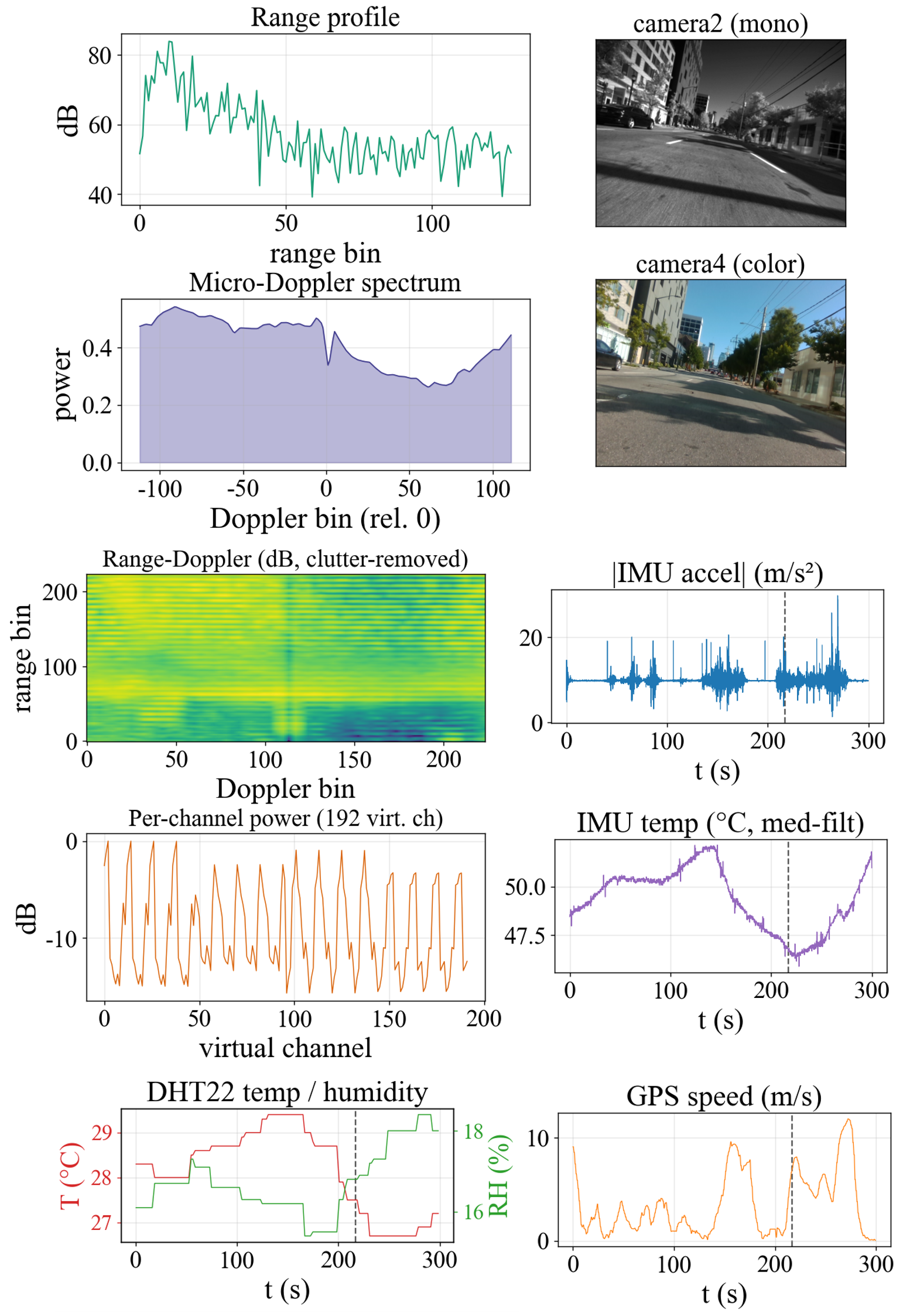}
\caption{Rad-R acquisition dashboard: yaw2, radar frame 2122, snapshot at 217 s~\cite{radr}. Radar views, camera frames, and auxiliary-sensor traces illustrate the recorded context.}
\label{fig:panel}
\end{figure}

\subsection{Selective state spaces and set representations}
Mamba makes state transitions and readout depend on sequence content~\cite{mamba}. We apply a local selective-SSM block to the 64 radar loops associated with a selected range response within each frame. A shared peak encoder followed by mean pooling follows the permutation-invariant construction of Deep Sets~\cite{deepsets}. VICReg supplies agreement, variance, and covariance penalties for source self-supervision~\cite{vicreg}.

Structured state-space models such as S4 provide a foundation for efficient sequence processing~\cite{s4}; Mamba-2 subsequently relates selective state spaces to structured attention through state-space duality~\cite{mamba2}. Masked autoencoders learn representations through reconstruction of missing image patches~\cite{mae}. SCORE corrupts response coordinates and reconstructs every valid token coordinate, combining this objective with split-set consistency. Its readout differs from episodically trained prototypical networks~\cite{protonet}: source class centers are fitted after self-supervision and compared as healthy-relative directions.

\subsection{Language interfaces for diagnosis}
BearLLM combines a vibration classifier, learned feature-to-word alignment, and language adaptation for bearing health management~\cite{bearllm}. Its multimodal dataset and bearing benchmarks differ from our radar recordings, preventing direct numerical comparison. We use the idea of separating diagnostic representation learning from language alignment, while measuring whether generated fault statements preserve the specialist's decisions.

Low-rank adaptation updates small trainable factors inside a frozen language model~\cite{lora}. Visual instruction tuning also illustrates learned projection into a language embedding space~\cite{llava}. In \lm{}, four fault similarities condition a local language model for maintenance guidance.

FaultGPT studies vibration-based fault-diagnosis question answering through time--frequency image/text instruction pairs and a multi-scale cross-modal image decoder~\cite{faultgpt}. FD-LLM also investigates language models for complex-equipment fault diagnosis~\cite{fdllm}. SCORE-LM uses a compact diagnostic-score interface for radar maintenance guidance.

Prefix-tuning optimizes continuous conditioning vectors with a frozen language base~\cite{prefix}; BLIP-2 learns a querying module to connect frozen visual and language models~\cite{blip2}. Our five input-dependent tokens follow the continuous-conditioning approach. For evaluation, FActScore distinguishes supported and unsupported atomic statements within long answers~\cite{factscore}. We separately measure fault correctness and answer-level unsupported observations.

Hallucination research distinguishes fluent generation from faithful use of its evidence~\cite{hallucination}. SelfCheckGPT uses consistency across sampled responses to detect potential hallucinations~\cite{selfcheck}; our review checks saved assertions against the supplied information. Retrieval-augmented generation provides external documents~\cite{rag}, whereas SCORE-LM conditions generation on diagnostic scores and a question.

\section{Rad-R Acquisition and Evaluation Data}
\subsection{Capture setup}
We use the Rad-R dataset~\cite{radr}. The system employs a 77-GHz TI MMWCAS-RF-EVM cascade based on four AWR1243 devices, with 12 transmit slots and 16 receivers, yielding 192 virtual channels. Each evaluated complex frame contains 64 loops and 256 ADC samples per chirp. The acquisition records 8-Msps ADC data with a nominal 100-ms frame period. Figure~\ref{fig:rig} shows the vehicle-mounted hardware and assembled sensor plate.

The recordings contain one healthy condition and two acquisition settings for each of four faults. Vibration is induced at 20/40 Hz; yaw misalignment at 5/10 degrees; blockage covers approximately 30/60\% of the radome; receiver degradation uses conductive foil affecting 6/16 or 10/16 receivers. Recording identifiers follow the dataset naming convention.

Rad-R also provides synchronized camera, IMU, temperature, and positioning streams, illustrated in Fig.~\ref{fig:panel}. SCORE processes radar responses; one retained legacy descriptor additionally uses ego-speed availability in a threshold gate. The language interface receives the four resulting fault similarities.

\subsection{Cache and capture exclusions}
The experiments use an existing 1,800-frame IQ cache containing 200 sampled frames from each recording, rather than every frame in the full acquisition. Ten consecutive blockage frames with an all-zero receiving device are excluded from evaluation, leaving 1,790 eligible frames. Eight leave-one-fault-capture-out folds each withhold an entire fault recording. Its counterpart at the other acquisition setting remains source data. Normalization, representation fitting, and readout calibration exclude the target capture.

The single healthy recording is always a source recording. The benchmark tests cross-recording discrimination among four faults. Severity and recording identity vary together. Repeated seeds quantify fitting variability on these recordings.

\subsection{Observation units}
We form nonoverlapping decisions inside contiguous valid blocks. Windows never cross capture boundaries, excluded frames, or injected gaps. The 1/10/50/100-frame horizons provide 1,590/158/31/15 fault decisions per seed. At the cache's nominal sampling rate of $10/3$ Hz, these budgets correspond to approximately 0.3/3/15/30 s of observation. Larger budgets provide fewer decisions, making the 100-frame result particularly sensitive to individual errors.

\section{SCORE Radar Representation}
\begin{figure*}[!t]
\centering\includegraphics[width=\textwidth]{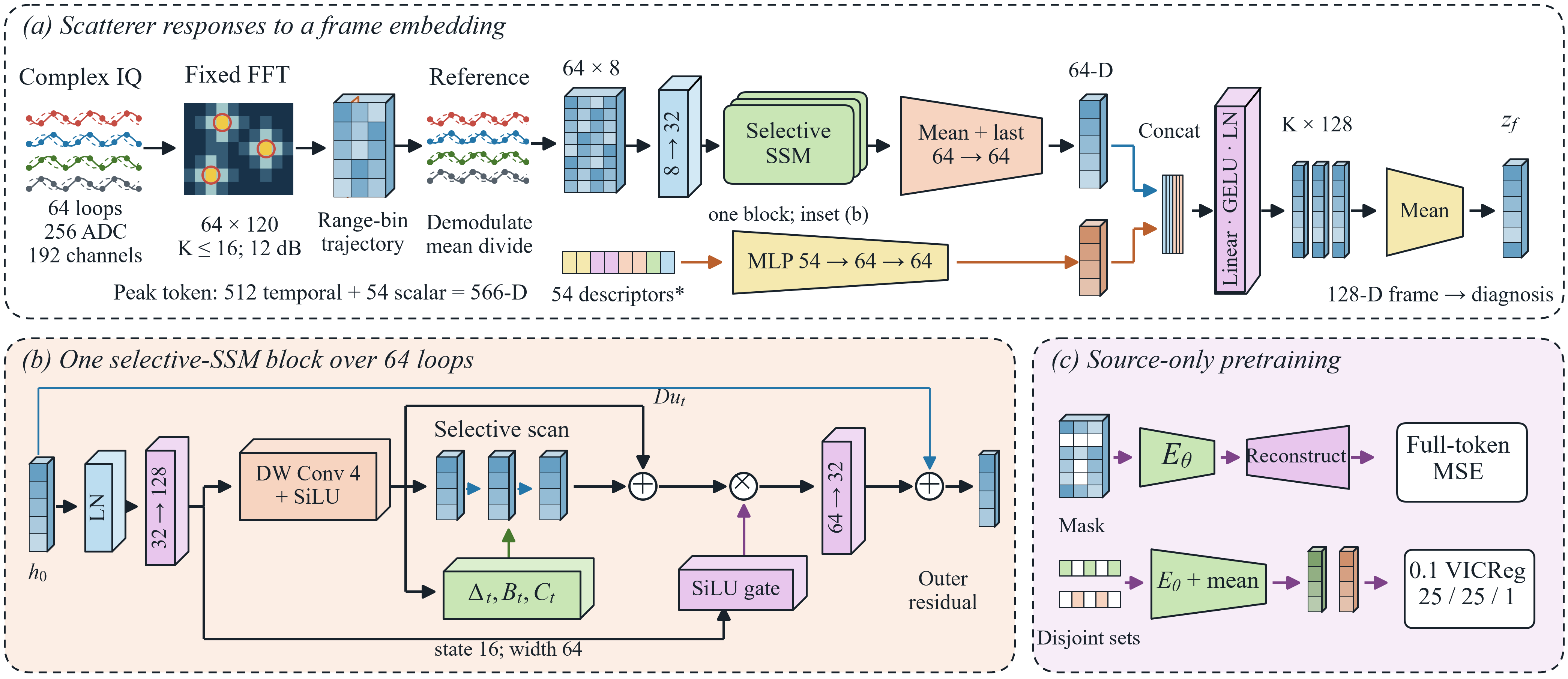}
\caption{Frozen SCORE signal pipeline, with the selective-state-space block and source self-supervision. Each peak contributes 512 temporal and 54 descriptor coordinates. Temporal and descriptor branches are fused \emph{before} mean pooling over peaks. Mean-plus-last pooling over radar loops is distinct from frame or peak averaging. The three legacy descriptors are uncalibrated heuristics. The diagram is schematic; reconstruction is training-only. The language connection in Fig.~\ref{fig:language} is added after the frozen diagnostic readout.}
\label{fig:score}
\end{figure*}

Figure~\ref{fig:score} summarizes the response extraction, shared encoder, and training-only objectives. The representation separates within-frame loop processing from aggregation over peaks and frames.
For target recording $g$, let $\mathcal S_g$ contain the source frames and $\mathcal T_g$ the excluded recording, with $\mathcal S_g\cap\mathcal T_g=\varnothing$. Preprocessing statistics, encoder fitting, and diagnostic reference directions depend only on $\mathcal S_g$. Within a frame, $k$ indexes peaks and $t$ indexes radar loops; neither denotes a frame in the subsequent observation window. A valid standardized peak token is $x_{f,k}=[\operatorname{vec}(Q_{f,k});p_{f,k}]$, with $Q_{f,k}\in\mathbb R^{64\times8}$ and $p_{f,k}\in\mathbb R^{54}$. The same encoder parameters are shared by all peaks. Fold subscripts are suppressed below for readability.
\subsection{Scatterer-conditioned tokens}
For each frame, fixed processing subtracts the ADC mean, computes a range FFT, and retains bins 8--127. A Doppler FFT over loops and channel-averaged magnitude form a $64\times120$ detection map. A $3\times3$ local-maximum detector selects peaks more than 12 dB above the map's fifth-percentile level, retaining at most 16. If no peak passes the threshold, it falls back to local maxima. The selected locations determine the response trajectories supplied to the encoder.

For a selected peak $k$, all 64 loop samples at its range bin are retained. We demodulate its Doppler component and coherently sum the 48 virtual channels associated with each receiving device. With demodulated device response $a_{k,t,d}$, self-referencing gives
\begin{equation}
q_{k,t,d}=\frac{a_{k,t,d}}{64^{-1}\sum_{\tau=1}^{64}a_{k,\tau,d}+\epsilon}.
\label{eq:response}
\end{equation}
Four complete complex device trajectories at the selected range bin yield 512 real coordinates, arranged as a $64\times8$ sequence.

An additional 54 coordinates comprise nine range-response samples, nine Doppler-response samples, 24 within-device receiver-phasor coordinates, four sorted relative device amplitudes, one peak-contrast value, four sorted device SNR descriptors, and three legacy geometry heuristics. The final three describe scene-relative Doppler extent, an adjacent-receiver phase summary, and their mismatch. These are uncalibrated scene-relative heuristics retained in both models of the branch-removal comparison.

\subsection{Shared encoder}
Coordinate means and scales are fitted on valid source tokens. The temporal sequence is projected from width eight to width 32, then processed by one pre-normalized residual selective-SSM block. The block expands to width 64, uses a causal depthwise convolution of width four, and maintains 16 state coordinates per inner channel. Its implemented recurrence is
\begin{align}
s_{t,i,n}&=e^{\Delta_{t,i}A_{i,n}}s_{t-1,i,n}
             +\Delta_{t,i}B_{t,n}u_{t,i},\\
y_{t,i}&=\sum_{n=1}^{16}C_{t,n}s_{t,i,n}+D_i u_{t,i},
\end{align}
where $i=1,\ldots,64$ indexes inner channels, $n=1,\ldots,16$ indexes state coordinates, and $s_{0,i,n}=0$. Here $u_t$ is the convolved and activated content stream, $A=-\exp(A_{\log})$, and $\Delta_t$ is positive and input-dependent. The vectors $B_t,C_t\in\mathbb R^{16}$ are input-dependent and shared across inner channels. A separate SiLU gate multiplies the content output before projection and the outer residual. Concatenated mean and final loop states are projected to a 64-dimensional temporal embedding.

The descriptor branch applies a $54\!\rightarrow\!64\!\rightarrow\!64$ MLP with GELU. Concatenated branch outputs pass through a $128\!\rightarrow\!128$ projection, GELU, and LayerNorm. A masked uniform mean across valid peaks gives frame embedding $z_f\in\mathbb R^{128}$. The nonlinear descriptor transform precedes peak averaging. The deployed encoder has 38,880 learned parameters.

Writing the shared temporal, descriptor, and fusion maps as $T_\eta,F_\eta,G_\eta$, respectively, the aggregation is
\begin{align}
e_{f,k}&=G_\eta\bigl([T_\eta(Q_{f,k});F_\eta(p_{f,k})]\bigr),\\
z_f&=\frac{\sum_{k=1}^{16}m_{f,k}e_{f,k}}{\max(1,\sum_{k=1}^{16}m_{f,k})},
\qquad \bar z_w=\frac{1}{|w|}\sum_{f\in w}z_f,
\label{eq:pooling}
\end{align}
where $m_{f,k}\in\{0,1\}$ identifies valid peaks. Equation~\ref{eq:pooling} is invariant to a permutation of peak indices because the same map is applied before a commutative sum. In particular, generally $F_\eta(\operatorname{mean}_k p_{f,k})\ne\operatorname{mean}_k F_\eta(p_{f,k})$; moving the nonlinear map after pooling would change the implemented model.

\subsection{Source-only pretraining}
Pretraining masks 25\% of loop positions, one device, and 20\% of descriptor coordinates after standardization. A training-only $128\!\rightarrow\!256\!\rightarrow\!566$ reconstruction head predicts the original token. Reconstruction MSE covers all coordinates of each valid token. In a second path, valid peaks are randomly divided into two disjoint subsets. The shared encoder and set mean yield paired embeddings $z_A,z_B$. Frames with an empty subset are excluded from this term; fewer than two valid pairs in a batch give zero consistency loss.

The objective is
\begin{equation}
\mathcal L_{\rm SSL}=\mathcal L_{\rm rec}+0.1(25\mathcal L_{\rm inv}+25\mathcal L_{\rm var}+\mathcal L_{\rm cov}).
\end{equation}
For $N_v$ valid tokens in a batch, corruption $\mathcal M$, and reconstruction head $R_\omega$, the implemented reconstruction term is
\begin{equation}
\mathcal L_{\rm rec}=\frac{1}{566N_v}\sum_{j=1}^{N_v}
\|R_\omega(E_\eta(\mathcal M(x_j)))-x_j\|_2^2,
\end{equation}
where $E_\eta$ denotes the shared peak encoder. The target is the uncorrupted standardized token; the denominator includes unmasked coordinates. For $N\geq2$ valid paired subset embeddings, stack the two views into $Z^A,Z^B\in\mathbb R^{N\times d}$, with $d=128$. Define the column-centered matrix $Z_c$ and sample covariance $C(Z)=Z_c^{\mathsf T}Z_c/(N-1)$. Then
\begin{align}
\mathcal L_{\rm inv}&=\frac{\|Z^A-Z^B\|_F^2}{Nd},\\
V(Z)&=\frac{1}{d}\sum_{j=1}^{d}
\left[1-\sqrt{C(Z)_{jj}+10^{-4}}\right]_+,\\
K(Z)&=\frac{1}{d}\sum_{i\ne j}C(Z)_{ij}^2,\\
\mathcal L_{\rm var}&=V(Z^A)+V(Z^B),\qquad
\mathcal L_{\rm cov}=K(Z^A)+K(Z^B),
\end{align}
where $[a]_+=\max(0,a)$. The two-view penalties are summed and encourage agreement without collapse or redundant coordinates~\cite{vicreg}. Disjoint peak indices can also share a range-bin trajectory.

Training uses 60 epochs, batch size 64, and AdamW with cosine scheduling. SSM parameters use learning rate $5\times10^{-4}$ and zero weight decay; other parameters use $10^{-3}$ and weight decay $10^{-4}$. Fault labels enter the subsequent readout, not this representation objective. The pretraining target capture exclusion is maintained throughout.

\subsection{Directional diagnostic readout}
For source ten-frame windows of class $c$, write $\mathcal W_c$ and define
\begin{equation}
\mu_c=\frac{1}{|\mathcal W_c|}\sum_{w\in\mathcal W_c}\bar z_w,
\qquad d_c=\frac{\mu_c-\mu_0}{\|\mu_c-\mu_0\|_2+\epsilon},
\end{equation}
where $c=1,\ldots,4$ denotes a fault and $\mu_0$ is the source healthy center. Source displacements are averaged before direction normalization. With $\epsilon=10^{-9}$, a query window gives
\begin{equation}
v_w=\frac{\bar z_w-\mu_0}{\|\bar z_w-\mu_0\|_2+\epsilon},\qquad
r_{w,c}=d_c^{\mathsf T}v_w.
\label{eq:scores}
\end{equation}
At every observation length, the decision is the fault with maximal similarity; there is no Healthy output. The four outputs are uncalibrated cosine similarities~\cite{calibration}. For nonzero displacements, ignoring the numerical $\epsilon$, multiplying a query's healthy-relative vector by any positive scalar leaves its scores unchanged. This gives radial invariance in the learned latent space. With fixed source directions and $\|v'-v\|_2\leq\delta$, Cauchy--Schwarz gives $|d_c^{\mathsf T}(v'-v)|\leq\delta$. Consequently a top-two score margin above $2\delta$ preserves the predicted class.

The same four-fault rule is used at 1, 10, 50, and 100 frames. A query averages its frame embeddings before subtracting the healthy center and normalizing. Source centers are always fitted from ten-frame windows. Label-efficiency experiments change only the selected source windows used to fit those centers. Frame-gap and selective-prediction diagnostics use the same directional decoder.

\section{Score-to-Language Alignment}
\begin{figure*}[!t]
\centering\includegraphics[width=\textwidth]{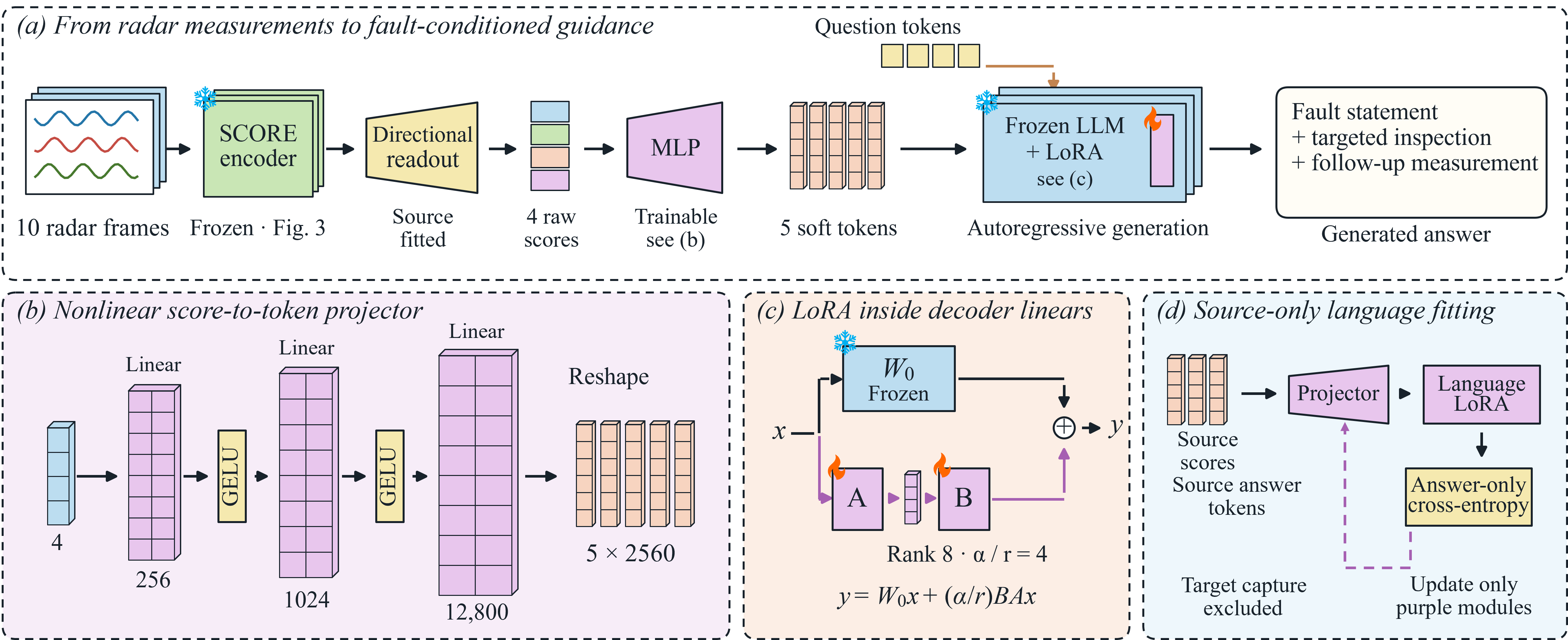}
\caption{Visual overview of score-to-language alignment. (a) Frozen SCORE and its directional readout produce four raw similarities; projected tokens and a question condition generation. (b) Three-layer nonlinear projector, with widths shown for Qwen. (c) Low-rank updates inside frozen decoder linears. (d) Source-only answer-token training updates the projector and LoRA, not SCORE or base-language weights. Tensor shapes and traces are schematic; no textual fault label is supplied as input.}
\label{fig:language}
\end{figure*}
\subsection{Interface and trainable modules}
\lm{} freezes SCORE and its source-fitted directional readout. The ordered vector
\begin{equation}
r_w=[r_{\rm vibration},r_{\rm misalignment},r_{\rm blockage},r_{\rm degradation}]
\end{equation}
is the sole radar-derived input to the language projector. No textual predicted fault, reference answer, or measurement packet is inserted into the question prompt. The selected projector is
\begin{equation}
P_\theta(r)=W_3\,\mathrm{GELU}(W_2\,\mathrm{GELU}(W_1r+b_1)+b_2)+b_3,
\end{equation}
with widths $4\!\rightarrow\!256\!\rightarrow\!1024\!\rightarrow\!12800$. Its output is reshaped into five 2560-dimensional soft tokens for Qwen3.5-4B~\cite{qwen}. The projector receives the raw cosine similarities directly.

The MLP contains 13,384,448 trainable parameters. Current maintenance training uses rank-eight language LoRA, scaling parameter 32, and dropout 0.1, adding 16,232,448 trainable language parameters. For a selected linear operation, LoRA represents an update as $\Delta W=(\alpha/r)BA$~\cite{lora}. The resulting trainable alignment stage has 29,616,896 parameters, excluding the frozen multi-billion-parameter language backbone.

The score bottleneck supplies class-level diagnostic information. Measurement-specific explanations require additional inputs such as noise floor, Doppler oscillation, phase variance, or receiver-resolved observations.
More precisely, for fixed trained weights and question $q$, any two radar windows satisfying $r(x)=r(x')$ induce the same conditional answer distribution:
\begin{equation}
p_{\theta,\phi}(a\mid x,q)=p_{\theta,\phi}(a\mid x',q).
\end{equation}
This follows directly because the generator receives only $P_\theta(r)$ and the text. Nonlinearity increases the projector's representational flexibility but cannot distinguish evidence collapsed by the score map.

\subsection{Training questions and targets}
For each target-capture fold, the current maintenance set selects ten source windows per annotated fault, giving 40 windows. Six question wordings per window yield 240 training pairs per fold. Across eight separately trained folds this gives 1,920 fold-specific pairs. Source windows can recur across folds, with each fold excluding its own target recording.

Targets distill the source SCORE decision into a fault statement and class-level inspection guidance. A selected source window in the yaw2 fold has an annotation/teacher disagreement: its annotated class is misalignment but SCORE predicts vibration. The retained target expresses the specialist's prediction.

The guidance has three steps: inspect the relevant installation or sensing path; make a conditional correction using the equipment procedure; and repeat a controlled measurement. These assistant-authored templates were reviewed by the user for task scope; independent expert maintenance validation remains pending. All frame sets corresponding to the 24 language-evaluation windows are excluded from generated language-training inputs. Historical SCORE pretraining retains its original source-only exclusions.

\subsection{Optimization and decoding}
The MLP and language adapters train jointly with answer-token cross-entropy,
\begin{equation}
\mathcal L_{\rm QA}=-\frac{1}{|\mathcal A|}\sum_{j\in\mathcal A}\log p_{\theta,\phi}(a_j\mid P_\theta(r_w),q,a_{<j}),
\end{equation}
where $q$ is the question, $a_j$ the answer token, $\theta$ the projector parameters, $\phi$ the language-adapter parameters, and $\mathcal A$ the supervised answer positions. Prompt tokens do not contribute to the objective. Each microbatch averages over its answer positions; four microbatch losses are averaged for an update. The language base, SCORE encoder, and readout stay frozen. Each fold uses two epochs, batch size one, gradient accumulation four, and 120 optimizer updates. AdamW~\cite{adamw} uses learning rate $5\times10^{-5}$, weight decay 0.01, cosine scheduling, and gradient clipping at one. Seed zero and final-only checkpoint selection are fixed. Input/loss checks, finite gradient and update checks, frozen-base checks, and final checkpoint reload checks accompany training.

All four language backbones use greedy generation, a maximum of 384 new tokens, and disabled explicit reasoning output. Capped answers are retained and identified. The backbones share the maintenance data, fold routing, final-checkpoint selection and optimization budget; each uses its native chat template and tokenizer.

\section{Experimental Design}
\subsection{Classifier competitors}
The primary comparison retains five seeds for every method. RD-CNN ERM, adapted CORAL, and MixStyle use an independent four-device RD-map CNN with 75,157 parameters. DARM uses a distance-based prototype head with 75,152 parameters. Their primary RD recipe uses 60 epochs and class-balanced supervision. CORAL excludes Healthy from covariance alignment. MixStyle applies random-pair mixing with probability 0.5 and Beta parameter 0.1 in the first two CNN blocks, and is inactive at inference. DARM retains its three distance-aware losses and source-capture domain identities.

DDDG uses a radar-adapted CNN with RandMix, dual disentanglement, and a recurrent Gumbel mask, trained for 100 epochs. Its training graph has 1,214,111 parameters, while fault inference uses 141,141. RadrNet-DS-CI uses independent IQ and RD branches, adaptive fusion, and fault/severity heads. Its primary recipe uses official CUDA Mamba, 20 epochs, batch size 16, AdamW learning rate and weight decay $10^{-4}$, cosine scheduling, and gradient clipping at one. Fault CE plus 0.5 times severity CE gives it auxiliary labels not used by SCORE. Checkpoints are fixed final epochs, not chosen by target accuracy.

BDC and DWCN use four source-standardized RD channels flattened to sequences of length 7,680. BDC retains its official five-stage CNNs, learned adversarial perturbation, covariance consistency, cross-whitening, and supervised contrastive objectives. It uses Adam at $5\times10^{-4}$, weight decay $5\times10^{-5}$, and batch size 40. DWCN retains its db4 wavelet branches, cross-contrast perturbation, and class/instance consistency objectives, with Adam at 0.01, zero weight decay, and batch size 64. Both use FP32, 100 epochs, eight folds and five seeds, and fixed final checkpoints. Their deployed graphs contain 1,034,869 and 268,181 parameters, respectively. These adaptations preserve the upstream learning mechanisms while adjusting input channels and length-dependent layers for radar.

The primary readouts differ at every horizon: SCORE cannot output Healthy, whereas competitors retain native five-class predictions. A Healthy prediction on a fault example counts as an error. The systems also differ in pretraining and auxiliary-label access as specified above.

\begin{figure}[!t]\centering\includegraphics[width=\columnwidth]{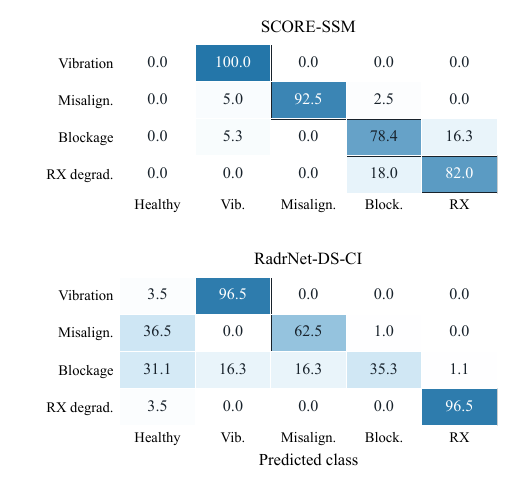}
\caption{Ten-frame confusion matrices using a common row-percentage scale and five seeds. Each seed contributes 40 vibration, 40 misalignment, 38 blockage, and 40 degradation decisions. There is no true Healthy row; SCORE's empty Healthy prediction column follows its four-fault directional readout.}
\label{fig:confusion}\end{figure}

\subsection{Label budgets and component controls}
In the low-label experiment, one label supervises a ten-frame window rather than ten frame losses. The same stored selections are used for budgets 1, 2, 5, 10, 20, and all available source windows per class, with five fixed resamples. SCORE adds unlabeled source pretraining before source-fitted directional centers; neural competitors train on selected labeled inputs. RadrNet uses one intact ten-frame window per update because a larger window batch exceeded memory. This changes BatchNorm grouping and update counts relative to some competitors. Each method retains its registered training recipe. BDC and DWCN use batches of 16 windows for 100 epochs with final checkpoints. Their classification losses act on window-mean logits; BDC retains per-frame covariance and whitening losses, and DWCN retains its per-frame perturbation loss.

The branch-removal experiment excludes the entire IQ/SSM path while retaining the nonlinear $54\!\to\!64\!\to\!64$ descriptor MLP before peak averaging. Its physical fusion maps 64 to 128 dimensions, and its reconstruction head predicts only the 54 retained coordinates. Both variants use the primary directional readout, all eligible source ten-frame windows, eight held-out recordings, and five encoder seeds. The physical-only encoder trains for 60 source-only self-supervised epochs with final checkpoints. Source-frame selection and normalization match original SCORE pretraining, including frames excluded later by the readout quality mask. Shared-layer initialization and minibatch randomization are matched to the original recipe.

\subsection{Language evaluation and units of analysis}
The current language set contains 75 maintenance questions attached to 24 windows from eight fault recordings. These are previously exposed development cases. Repeated wordings test sensitivity to the question but do not create independent radar evidence. We report both all-answer counts and a canonical 24-window view using the smallest question ID for each window, never the best-performing wording.

Two distinct targets are retained: the recording's annotated fault and SCORE's frozen decision. True-fault accuracy measures agreement with the annotation; decoder agreement measures faithful transmission of the specialist. An LLM may improve the former by disagreeing with a mistaken specialist, or reduce it by changing a correct prediction. Neither quantity alone measures maintenance quality.

For the no-LoRA control, we remove language LoRA while keeping the \emph{same jointly trained projector}, score vectors, questions, and decoding. This is a component-removal intervention, not a separately optimized projector-only model. Fault judgments include full-answer contradictions, not merely the first fault word. Unsupported-evidence review counts assertions of observed physical details not established by the four-score input. General descriptions and conditional inspection suggestions are not counted solely for being detailed. Review is post-hoc and assistant-based, not independent expert adjudication.

\subsection{Metrics and uncertainty}
The primary classifier metric averages recall equally across the eight captures. Four-fault balanced accuracy and macro-F1 are additionally pooled over decisions. Population SD across five seeds describes fitting variation. The Full/Physical-only comparison uses the same five seeds; the retained earlier input controls use three seeds and sample SD. Low-label SD is across five resample means after averaging SCORE encoder seeds. These quantities are not confidence intervals over new recording sessions.

Language results use explicit numerators and denominators rather than inferential tests treating 75 questions as independent signals. We separately report truncation, contradictions, and unsupported observational claims; overlapping error categories are not summed. All records preserve raw outputs and data/checkpoint identities. Verification checks their consistency.

For seed $s$ and $G=8$ captures, each with $n_g$ decisions, capture recall is
\begin{equation}
R_s=\frac{1}{G}\sum_{g=1}^{G}\frac{1}{n_g}\sum_{i=1}^{n_g}
\mathbf1[\hat y_{s,g,i}=y_g].
\end{equation}
For pooled counts over the same decisions, four-fault macro-F1 is $\frac14\sum_{c=1}^{4}2TP_c/(2TP_c+FP_c+FN_c)$. A Healthy prediction contributes a false negative for the true fault. These metrics weight examples differently when capture sizes differ.

For $M=75$ language answers, let $\ell_j\in\{1,2,3,4,\bot\}$ be the full-answer fault judgment, with $\bot$ for unresolved or contradictory diagnoses, $y_j$ the annotation, and $\tilde y_j$ the frozen specialist's decision. We compute
\begin{align}
A_{\rm fault}&=M^{-1}\sum_j\mathbf1[\ell_j=y_j],\\
A_{\rm agree}&=M^{-1}\sum_j\mathbf1[\ell_j=\tilde y_j].
\end{align}
For either metric, canonical-window evaluation restricts the sum to the fixed 24 question IDs and replaces $M$ by 24. If $u_j$ flags clear unsupported observations and $e_j=\mathbf1[\ell_j\ne y_j]$, the combined error flag is $e_j+u_j-e_ju_j$. This prevents double-counting an answer that is both diagnostically wrong and unsupported.

\begin{figure}[!t]\centering\includegraphics[width=\columnwidth]{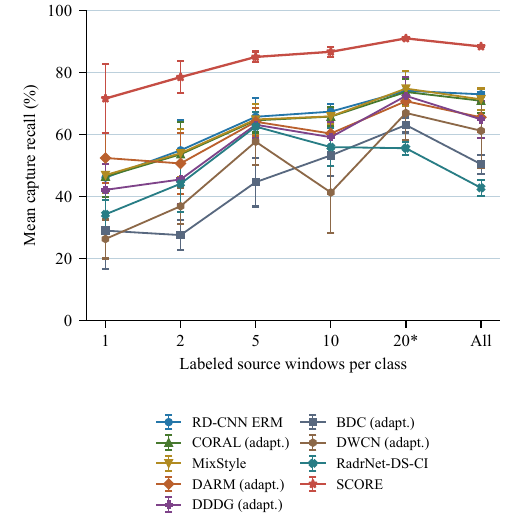}
\caption{Nine-method low-label transfer with common source-window selections. Error bars are population SD across five resample means. SCORE additionally uses unlabeled source pretraining. One label is one ten-frame window. The 20-window budget uses seven folds because one fold has only 18 source blockage windows; other budgets use eight.}
\label{fig:labels}\end{figure}

\section{Radar Classification Results}
\subsection{Fault transfer and observation budget}
Table~\ref{tab:primary} reports the primary comparison. SCORE reaches 88.39\% ten-frame mean capture recall, 15.56 percentage points above the highest comparator mean, RadrNet-DS-CI at 72.83\%. Corresponding macro-F1 values are 88.20\% and 77.37\%.

\begin{table*}[t]\centering\footnotesize
\caption{Primary capture-mean recall by frame budget and pooled ten-frame balanced accuracy (BA) and macro-F1, in percent. All five seeds retained. Adapted methods use radar-specific implementations.}
\label{tab:primary}\setlength{\tabcolsep}{6pt}
\begin{tabular}{lrrrrrr}\toprule
Method & 1 frame & 10 frames & 50 frames & 100 frames & 10-frame BA & 10-frame F1\\\midrule
RD-CNN ERM &69.68&71.94&75.62&80.00&72.12&71.36\\
CORAL (adapted)&68.94&70.18&72.50&73.75&70.29&68.85\\
MixStyle&61.99&63.29&65.00&73.75&63.34&63.45\\
DARM (adapted)&61.31&64.62&68.75&73.75&64.68&63.22\\
DDDG (adapted)&67.51&69.40&70.83&71.25&69.39&69.27\\
BDC (adapted)&59.00&63.69&67.71&67.50&63.72&64.48\\
DWCN (adapted)&61.93&62.89&66.25&68.75&62.88&61.24\\
RadrNet-DS-CI&71.00&72.83&79.38&85.00&72.69&77.37\\
SCORE&\textbf{84.56}&\textbf{88.39}&\textbf{88.54}&\textbf{96.25}&\textbf{88.23}&\textbf{88.20}\\\bottomrule
\end{tabular}\end{table*}

Table~\ref{tab:primary} reports SCORE recalls of 84.56/88.39/88.54/96.25\% at 1/10/50/100 frames under the same directional decision rule. Longer horizons change the recording partitions and reduce the number of decisions. The 100-frame budget contains 15 decisions per seed, compared with 158 at ten frames.

The class breakdown in Fig.~\ref{fig:confusion} shows that the aggregate gain is not uniform. SCORE's pooled recalls are 100\% for vibration, 92.5\% for misalignment, 78.4\% for blockage, and 82.0\% for degradation. RadrNet achieves 96.5\% degradation recall, exceeding SCORE for this class. Confusion between blockage and degradation remains a central failure mode. An amplitude or response-quality change can support both labels, and the results alone do not identify which physical mechanism caused an error.

BDC and DWCN reach 63.69\% and 62.89\% ten-frame capture-mean recall, respectively. Their pooled vibration recalls are 90.5\% and 91.0\%, and misalignment recalls are 77.0\% and 84.0\%. Blockage transfer is substantially weaker: 17.89\% for BDC and 0.53\% for DWCN, compared with 78.4\% for SCORE. Their degradation recalls are 69.5\% and 76.0\%. Thus the new methods' aggregate shortfall is concentrated in blockage.

\subsection{Label efficiency}
SCORE's directional readout achieves 71.58/78.49/85.01/86.67\% mean capture recall with 1/2/5/10 labeled windows per class. At the smallest budget, DARM is the strongest selected competitor at 52.39\%, giving an observed gap of 19.19 percentage points. Figure~\ref{fig:labels} summarizes the nine-method low-label comparison. BDC reaches 28.93/27.49/44.58/53.22\%, and DWCN reaches 26.24/36.82/57.65/41.25\%, at the same four budgets.

Source self-supervision reduces labeled-data demand. RadrNet reaches 42.74\% with all eligible windows under its one-window-update recipe. The 20-window budget uses seven folds, while the all-window budget uses eight.

\subsection{Representation visualization}
\begin{figure}[!t]\centering\includegraphics[width=\columnwidth]{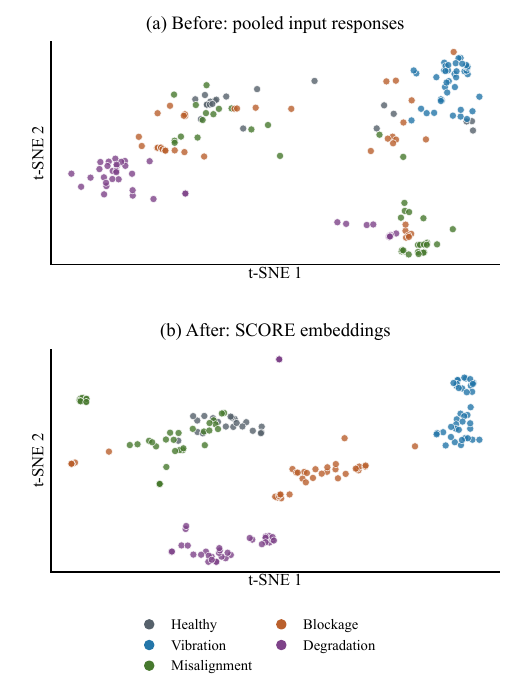}
\caption{Before and after SCORE encoding for 178 matched ten-frame windows: 160 source windows and 18 held-out Blockage1 windows. Healthy samples are source-only. The inputs have 566 coordinates and the embeddings 128. Both t-SNE fits use cosine distance, perplexity 30, 1,000 iterations, and identical random initialization; labels only set plot colors. Each panel has its own two-dimensional coordinates.}
\label{fig:tsne}\end{figure}

Figure~\ref{fig:tsne} compares the same 178 ten-frame windows before and after SCORE encoding using t-SNE~\cite{tsne}. Before encoding, standardized input responses are averaged over valid peaks and frames; after encoding, the learned frame embeddings are averaged over the same windows. We fix the Blockage1-excluded encoder and seed zero. The encoded view groups most blockage windows together and separates most degradation windows, while healthy and misalignment windows still overlap. These two-dimensional neighborhoods provide a qualitative view alongside the classification results.

\subsection{IQ/SSM branch removal}
Table~\ref{tab:ablation} reports the four component conditions. Full SCORE and the descriptor-only encoder use the primary directional protocol. Mean recording recall is 88.39\% for full SCORE and 87.47\% after removing the IQ/SSM branch, a difference of 0.92 percentage points.

\begin{table}[t]\centering\footnotesize
\caption{Component ablations: mean recall and macro-F1 (\%). Full and Physical only use the five-seed directional protocol (population SD). The last two rows retain the earlier three-seed input-ablation protocol (sample SD).}
\label{tab:ablation}\setlength{\tabcolsep}{3pt}
\begin{tabular}{lrrr}\toprule
Input&1-frame recall&10-frame recall&10-frame F1\\\midrule
Full&84.56 (1.09)&\textbf{88.39} (0.70)&\textbf{88.20} (0.67)\\
Physical only&\textbf{84.62} (1.60)&87.47 (1.47)&87.29 (1.37)\\
Temporal only&27.45 (1.17)&44.58 (4.85)&39.61 (5.32)\\
No legacy geometry&74.34 (3.24)&86.11 (2.25)&85.86 (2.26)\\\bottomrule
\end{tabular}\end{table}

Four-fault macro-F1 is 88.20\% versus 87.29\%. The descriptor-only encoder contains 16,256 parameters compared with 38,880 in full SCORE. Independent functional checkpoint replay reproduces the branch-removed embeddings, directional scores, and predictions for all 40 fits.

Both variants encode each peak nonlinearly before averaging, preserving the descriptor branch's within-frame processing. Removing the IQ/SSM branch also reduces the fusion width and reconstruction target. This matched intervention measures the contribution of the complete branch and its fusion. Temporal-only and no-geometry rows retain the earlier quality-filtered pretraining and one-frame linear readout; they are not recomputed under the new protocol.

\subsection{Processing cost and saved-output diagnostics}
\begin{figure}[!t]\centering\includegraphics[width=\columnwidth]{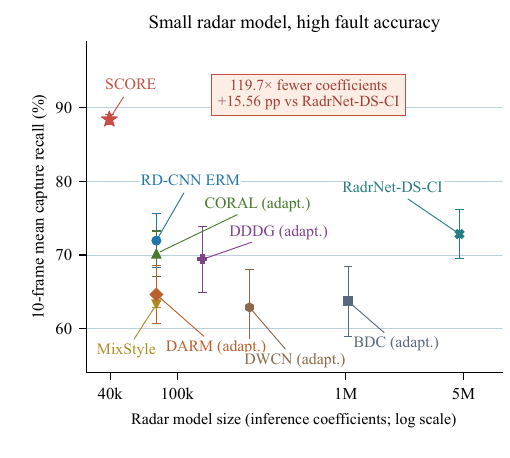}
\caption{Radar model size versus ten-frame fault accuracy, measured as mean capture recall over eight held-out recordings. Error bars: population SD across five seeds. Size counts active scalar inference weights and fitted readout coefficients; scaler statistics, running buffers, training-only heads, and all language modules are excluded. SCORE includes its 640 center/direction values; RadrNet excludes its unused severity head. Adapted methods and readout/supervision differences follow Sec.~VI. The highlighted gain is relative to RadrNet-DS-CI, the highest-recall competitor.}
\label{fig:size}\end{figure}

SCORE's primary deployed graph contains 38,880 encoder parameters plus 640 center/direction coefficients, totaling 39,520. RadrNet's fault graph contains 4,731,850 coefficients after excluding the unused severity head; DDDG uses 141,141. Figure~\ref{fig:size} shows the accuracy--size trade-off for all nine methods. SCORE uses 119.7$\times$ fewer inference coefficients than RadrNet-DS-CI while gaining 15.56 percentage points in mean recall. It is the only nondominated mean among these nine radar classifiers. Table~\ref{tab:timing} reports processing measurements on the same physical RTX A4000. SCORE's directional timing graph has a 92.86 ms composed median and a 9.97 ms GPU-only median. Its compact parameter count does not imply the fastest GPU forward.

\begin{table*}[t]\centering\footnotesize
\caption{Processing measurements from separate runs: batch one, one CPU thread, 20 warm-ups and 200 synchronized repeats. Composed timing is measured directly.}
\label{tab:timing}\setlength{\tabcolsep}{6pt}
\begin{tabular}{llrrr}\toprule
Runtime&Graph&GPU median / p99 (ms)&DSP median (ms)&DSP + GPU median / p99 (ms)\\\midrule
Windows&DDDG&\textbf{1.04} / \textbf{1.93}&132.83&138.13 / 156.87\\
WSL&RadrNet-DS-CI&8.51 / 14.68&89.18&147.82 / 197.58\\
WSL&SCORE directional graph&9.97 / 15.03&\textbf{79.69}&\textbf{92.86} / \textbf{122.19}\\\bottomrule
\end{tabular}\end{table*}

Timing uses one fixed healthy cached frame and excludes disk I/O. Historical primary SCORE weights were not saved, so its timing uses the unchanged encoder graph with architecture-equivalent random weights and source-fitted directional centers. The directional head agrees with an independent NumPy calculation within $1.2\times10^{-7}$. Processing ten cached frames together takes a median 8.63 ms on the GPU, excluding DSP. RadrNet composed timing recomputes RD inputs from quantized IQ rather than the original pre-quantization RD cache. These measurements cover radar processing; language generation is additional.

Directional saved-embedding diagnostics use source-fitted thresholds and fixed centers. A maximum-softmax threshold targeting 90\% source-fault coverage, with temperature one applied to the four cosine scores, gives 65.86\% mean per-recording target coverage at 98.96\% selective capture-mean recall. Accepted windows cover seven or eight recordings per seed; the selective mean averages only recordings with accepted windows. Central frame-gap removal rebuilds windows without bridging gaps; 0/5/10/25\% removal leaves 158/143/142/111 decisions per seed. SCORE obtains 88.32\% recall at a 10\% gap using the same directional decoder.

\section{Language Results}
\subsection{Projector and score representation}
The interface study first used short fault questions alongside maintenance and risk prompts. Table~\ref{tab:projector} reports only its 25 diagnosis questions, corresponding to 24 distinct radar windows. All variants used the same exposed development set. The raw-score full MLP achieved 23/25 true-fault answers and 24/25 agreement with the frozen decoder. This result selected the interface subsequently used for maintenance training.

\begin{table}[t]\centering\footnotesize
\caption{Historical projector development study, fixed 25 short diagnosis questions. The full MLP and linear projectors differ in capacity and initialization.}
\label{tab:projector}\setlength{\tabcolsep}{5pt}
\begin{tabular}{llrr}\toprule
Score transform&Projector&True fault&Decoder agreement\\\midrule
Softmax&Linear&20/25&22/25\\
Sigmoid&Linear&20/25&22/25\\
Raw&Linear&21/25&20/25\\
Sigmoid&Full MLP&6/25&5/25\\
Raw&Full MLP&\textbf{23/25}&\textbf{24/25}\\\bottomrule
\end{tabular}\end{table}

With sigmoid plus MLP, 22/25 diagnosis outputs selected misalignment despite decreasing training loss. Conversely, raw-score linear projection improved annotation accuracy without improving decoder agreement. The same development questions informed these successive interface choices.

\subsection{Maintenance generation with and without LoRA}
The current rank-eight maintenance model was trained on 40 source windows per fold and evaluated on 75 maintenance questions (Table~\ref{tab:qa}). All 75 adapted responses consist of one of four class-specific reference templates. Reading the four unique complete responses and checking their membership for every item confirms the leading-fault parser's labels and finds no additional asserted measurement evidence in these templates.

\begin{table}[t]\centering\footnotesize
\caption{Current maintenance evaluation. Same scores, questions, jointly trained projector and decoding; only language LoRA is removed in the control. Counts cover 75 questions on 24 radar windows.}
\label{tab:qa}\setlength{\tabcolsep}{3pt}
\begin{tabular}{lrr}\toprule
Measure&With rank-8 LoRA&LoRA removed\\\midrule
Correct fault, all questions&\textbf{62/75 (82.7\%)}&45/75 (60.0\%)\\
Decoder agreement, all&\textbf{67/75 (89.3\%)}&47/75 (62.7\%)\\
Correct fault, canonical windows&\textbf{20/24}&17/24\\
Decoder agreement, canonical&\textbf{22/24}&18/24\\
Output cap reached&\textbf{0/75}&20/75\\
Unsupported observations$^\dagger$&\textbf{0/75}&$\geq31/75$\\\bottomrule
\multicolumn{3}{p{.95\columnwidth}}{\scriptsize $^\dagger$Post-hoc assistant review, at answer level, of unsupported observations beyond the fault label. The adapted set reproduces four guidance templates.}
\end{tabular}\end{table}

Removing language adapters decreases correct fault statements by 17/75 answers and decoder agreement by 20/75. At the canonical-window level, differences are three correct faults and four decoder agreements. The control removes a co-adapted component from the jointly optimized projector--LoRA system.

The control's per-class correct counts are 8/19 vibration, 6/19 misalignment, 15/19 blockage, and 16/18 degradation. Ten of its 24 windows receive different fault diagnoses across question wordings. Two responses begin with vibration but later diagnose misalignment and are not credited as correct. Seventeen of the 20 truncated control answers nevertheless contain the correct fault statement, demonstrating why reaching the token cap and fault accuracy are separate outcomes.

\subsection{Unsupported detail versus useful elaboration}
\begin{figure}[!t]\centering\includegraphics[width=\columnwidth]{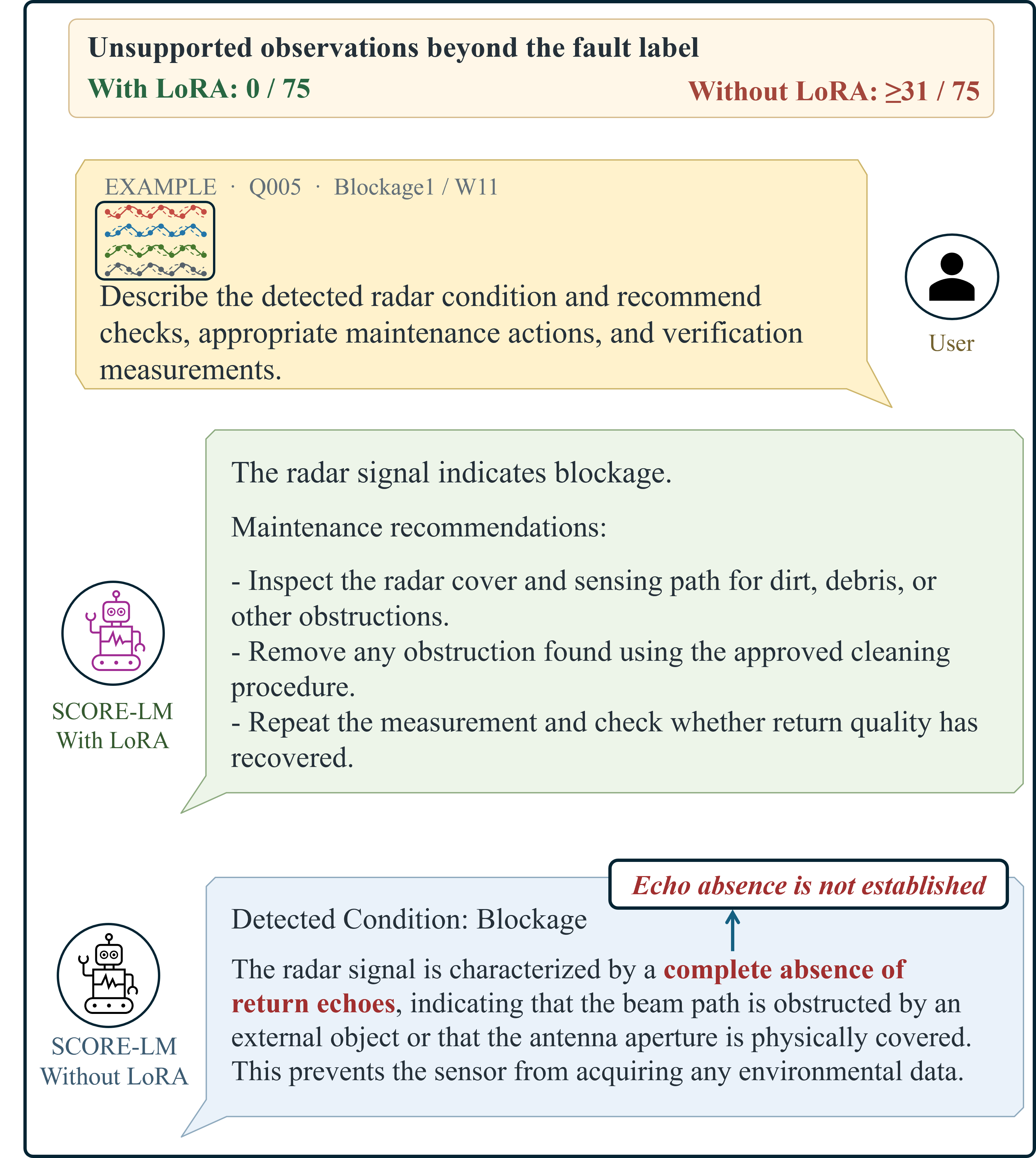}
\caption{Matched maintenance dialogue, Q005 (Blockage1, W11). Both models receive the same radar scores, question and jointly trained projector; the control removes language LoRA. The adapted answer is complete; the control is an exact excerpt. The signal icon is schematic. Highlighted boxes are post-hoc review annotations, not model outputs. Both fault labels are correct in this example, but the control asserts echo absence that the four-score input does not establish. Aggregate unsupported-observation counts exclude fault-label errors: the adapted model still makes 13/75 incorrect fault statements.}
\label{fig:chat}\end{figure}

Full-text review identifies at least 31/75 control answers that assert observed evidence not established by the inputs. Examples include a complete absence of echoes, increased Doppler frequency, elliptical or figure-eight return patterns, and invented literal signal descriptors. Figure~\ref{fig:chat} shows a matched chat example and separates the saved model text from review annotations. The review distinguishes asserted observations from general explanations and conditional inspection suggestions.

Sixteen of the 45 correct-fault control answers also contain unsupported evidence. The union of a wrong/contradictory fault or clear unsupported evidence is therefore at least $30+16=46$ of 75 answers, or 61.3\%. The corresponding narrow unsupported-observation count is zero for the four adapted templates, but the adapted model still gives 13 incorrect fault answers. LoRA reduces unsupported elaboration while retaining some diagnostic errors.

An incorrect adapted response, Q004, states ``The radar signal indicates degradation'' for a blockage recording. Conversely, the no-LoRA Q055 answer interprets an ``elliptical'' return pattern without receiving a geometric return image. These cases illustrate distinct diagnostic and evidence-support errors.

The benefit currently supported is therefore more constrained output and better fault transmission in this component-removal comparison. The adapted answers do not vary maintenance details with calibrated severity, amplitude, or receiver localization. A comparison with deterministic class-conditioned rendering would help quantify the value of language generation.

\subsection{Additional language backbones}
The backbone study includes Qwen3.5-4B, Phi-4-mini-instruct, Ministral-3-3B-Instruct-2512, and Granite-4.2-3B~\cite{qwen,phi,ministral,granite}. Backbone-specific projector output widths follow each language embedding dimension. Each backbone uses eight fold-specific fits and the same 75 maintenance questions (Table~\ref{tab:backbones}).

\begin{table*}[t]\centering\footnotesize
\caption{Matched maintenance QA with rank-8 language adapters and trained projectors. Greedy decoding, 384-token cap; 75 questions on 24 windows. Time is mean generation latency in seconds per answer on an RTX A4000, excluding model loading.}
\label{tab:backbones}\setlength{\tabcolsep}{4pt}
\begin{tabular}{lccccc}\toprule
Backbone&Correct fault / 75&Agreement / 75&Unsupported / 75&Capped / 75&Time / answer\\\midrule
Qwen3.5-4B&62&67&\textbf{0}$^\dagger$&\textbf{0}&4.75\\
Phi-4-mini-instruct&59&66&\textbf{0}$^\dagger$&\textbf{0}&2.98\\
Ministral-3-3B-Instruct-2512&54&57&\textbf{0}$^\dagger$&\textbf{0}&\textbf{2.67}\\
Granite-4.2-3B&\textbf{65}&\textbf{70}&\textbf{0}$^\dagger$&\textbf{0}&3.88\\\bottomrule
\multicolumn{6}{p{.95\textwidth}}{\scriptsize $^\dagger$Answer-level assistant review of unsupported observations beyond the fault label.}
\end{tabular}\end{table*}


Granite gives the highest observed fault accuracy, 65/75 (86.7\%), followed by Qwen at 62/75 (82.7\%), Phi at 59/75 (78.7\%), and Ministral at 54/75 (72.0\%). Granite also has the highest decoder agreement, 70/75 (93.3\%). On the 24 canonical windows, correct-fault counts are 21, 20, 19, and 18, respectively. Phi classifies all 19 vibration questions correctly, but only 12/19 misalignment questions; Granite reaches 16/19 in both classes and 18/19 for blockage. These differences show that changing the language backbone changes fault transmission even with the same frozen radar scores.

All four backbones finish within the output cap. Full-text review finds no extra asserted signal observations under the same narrow rubric. Qwen, Granite, and Ministral reproduce four maintenance templates; Phi produces 11 distinct texts, including an unestablished ``connecting shaft'' in one inspection suggestion. This hardware assumption is recorded separately from asserted signal observations. Mean generation latency ranges from 2.67 s for Ministral to 4.75 s for Qwen. The comparison fixes training examples and updates, while native tokenization, projector width, and adapter parameter counts vary by backbone.

\section{Discussion, Limitations, and Reproducibility}
\subsection{What the radar results support}
The experiments support strong fault discrimination for the tested capture-excluded Rad-R setting and reduced labeled-window demand with source self-supervision. One healthy recording cannot provide independent-session false-alarm estimates, and SCORE's decision is restricted to four faults. Likewise, the recording/severity coupling prevents attribution of transfer to severity invariance alone. More sessions, devices, backgrounds, and healthy recordings are needed to establish deployment robustness.

The five-seed branch-removal comparison shows a modest 0.92-point mean gain from retaining the IQ/SSM path. Both variants preserve the nonlinear per-peak descriptor encoder, which remains sufficient for strong performance. The intervention characterizes the combined branch and fusion; a matched alternative temporal encoder would be needed to isolate the selective-SSM architecture itself. The legacy geometry heuristics and fixed detector settings remain possible sources of dataset dependence.

\subsection{What the language results support}
The language interface preserves a useful separation between the radar specialist and a local text generator, and its adapters affect both fault transmission and elaboration. However, the current training targets are compact class-conditioned templates. Reproducing them demonstrates alignment at a narrow bottleneck, not reconstruction of physical evidence. The input cannot support statements such as ``receiver three failed'' or ``the phase variance increased'' without an additional validated evidence channel.

The 75 questions are repeated development queries on only 24 windows. They were reused during design decisions, and there is no untouched paper-scale confirmation set. The rank-eight revision changed source-window coverage, update count, rank, and effective adapter scaling together; it is a system revision rather than a clean isolated rank ablation. The no-LoRA comparison is a removal control with a co-trained projector. Expert maintenance review, usefulness assessment, and a separately trained projector-only comparison remain future work.

\subsection{Scope for automation}
For a practical automated platform, the radar classifier and language assistant have different reliability requirements. A deterministic fault output can remain available without language generation, while any recommendation should be presented as an inspection suggestion subject to equipment procedures. The present interface addresses four-fault classification and inspection guidance. Severity estimation, simultaneous-fault diagnosis, remaining-life prediction, and closed-loop repair are future extensions. Rad-R's multi-sensor measurements could support richer evidence-grounded explanations.

The computational distinction is similarly important. SCORE is compact, but the full language system contains a large frozen backbone and approximately 29.6 million trainable alignment parameters for the current Qwen configuration. Paired radar timing does not include generation. A small diagnostic encoder and a local LLM may be deployed on different schedules or devices; end-to-end energy, latency, and memory guarantees have not been established by this study.

\subsection{Reproducibility boundaries}
The archive retains split indices, source-window selections, normalization identities, code versions, raw predictions, language responses, checkpoint identities, and verification records. Original SCORE and Rad-R sources are preserved separately from language experiments. Verification distinguishes stored-result checks from fresh neural replay. Some historical classifier runs lack saved weights; their reported outcomes are supported by saved predictions and experiment records, not newly reproduced training trajectories.

RadrNet's reduced-label implementation retains documented CUDA floating-point training variation. The directional SCORE evaluation reproduces the original 88.39\% ten-frame headline from saved embeddings, and the all-label endpoint and full-model ablation use exactly the same source centers, windows, and decisions.

\section{Conclusion}
\lm{} connects a compact response-centered radar classifier to an adapted local language model through a four-score bottleneck. SCORE leads the nine-method comparison at ten frames and retains strong performance with limited source labels. The Qwen adapter-removal study shows improved fault transmission and reduced unsupported observational elaboration with language adaptation. Across four backbones, matched maintenance-answer accuracy ranges from 72.0\% to 86.7\%, with class-conditioned guidance and no capped answers. Future work will assess maintenance usefulness through independent expert review and new recordings, and extend the score interface with validated physical evidence.

\end{document}